\documentclass{article}

\usepackage{PRIMEarxiv}

\usepackage[utf8]{inputenc} 
\usepackage[T1]{fontenc}    
\usepackage{hyperref}       
\usepackage{url}            
\usepackage{booktabs}       
\usepackage{amsfonts}       
\usepackage{nicefrac}       
\usepackage{microtype}      
\usepackage{lipsum}
\usepackage{fancyhdr}       
\usepackage{graphicx}       
\graphicspath{{media/}}     

\usepackage{dirtytalk}
\usepackage{booktabs}
\usepackage{listings}
\usepackage[normalem]{ulem}
\usepackage{enumitem}

\title{Double-Exponential Increases in Inference Energy: The Cost of the Race for Accuracy

}

\author{
  Zeyu Yang \\
  Department of Engineering Science \\
  University of Oxford \\
  Oxford, UK \\
  \texttt{zeyu.yang@eng.ox.ac.uk} \\
  \And
  Karel Ad\'{a}mek \\
  Department of Engineering Science \\
  University of Oxford \\
  Oxford, UK \\
  \texttt{karel.adamek@eng.ox.ac.uk} \\
  \And
  Wesley Armour \\
  Department of Engineering Science \\
  University of Oxford \\
  Oxford, UK \\
  \texttt{wes.armour@oerc.ox.ac.uk} \\
}

\begin{document}
\maketitle

\begin{abstract}
Deep learning models in computer vision have achieved significant success but pose increasing concerns about energy consumption and sustainability. Despite these concerns, there is a lack of comprehensive understanding of their energy efficiency during inference. In this study, we conduct a comprehensive analysis of the inference energy consumption of 1,200 ImageNet classification models—the largest evaluation of its kind to date. Our findings reveal a steep diminishing return in accuracy gains relative to the increase in energy usage, highlighting sustainability concerns in the pursuit of marginal improvements. We identify key factors contributing to energy consumption and demonstrate methods to improve energy efficiency. To promote more sustainable AI practices, we introduce an energy efficiency scoring system and develop an interactive web application that allows users to compare models based on accuracy and energy consumption. By providing extensive empirical data and practical tools, we aim to facilitate informed decision-making and encourage collaborative efforts in developing energy-efficient AI technologies.

\end{abstract}


\section{Introduction}
Over the past decade, AI has achieved remarkable capabilities across various fields. However, these accomplishments have come at the cost of significant computational demands. AI research has traditionally prioritized achieving the highest possible accuracy, often disregarding considerations of model size, complexity, and data requirements.

As the field matures and more AI products and services transition into commercial deployment, computational cost is becoming a major concern. Google reported that the energy consumption of machine learning (ML) workloads constituted 10–15\% of its total energy usage from 2019 to 2021, with training accounting for 40\% and inference for 60\%~\cite{patterson2022carbon}. Similarly, Meta observed a power capacity distribution of 10:20:70 among experimentation, training, and inference in their AI infrastructure~\cite{wu2022sustainable}.

Moreover, the electricity consumption of these tech giants has been rising steadily. Google's electricity usage increased by an average of 21\% per year over the past decade, growing from 3.7 TWh in 2013 to 25.3 TWh in 2023~\cite{google_env_report_2019, google_env_report_2024}. Meta's electricity consumption grew by an average of 32\% per year over the past five years, from 4.9 TWh in 2018 to 15.0 TWh in 2023~\cite{meta_env_report_2024}. Data centers globally were estimated to consume about 1\% of global electricity and contribute 0.3\% of greenhouse gas emissions in 2018~\cite{masanet2020recalibrating,jones2018stop}. Furthermore, widespread adoption of autonomous vehicles could require as much electricity as all current data centers combined~\cite{sudhakar2022data}.

The high energy consumption associated with AI leads to several negative consequences. Economically, it results in higher capital costs for purchasing computing hardware and increased operational expenses for electricity and cooling. Environmentally, it produces a large carbon footprint, exacerbating climate change. Additionally, the substantial computational demands impede further development and innovation in AI. The soaring costs of purchasing or renting computing power close the door to many researchers, leaving only large tech companies as major players in the field. This centralization contradicts the open-source ethos that has traditionally driven AI and software development. Moreover, high energy consumption hinders the deployment of AI in edge scenarios where battery life and thermal design power (TDP) are constrained~\cite{yang2022instinctive, pereira2023not}.

Despite these concerns, there is a lack of comprehensive understanding of the energy consumption during inference and the energy efficiency of different models. In this study, we address this gap by measuring the inference energy consumption of 1,200 ImageNet classification models—a scale that, to our knowledge, is orders of magnitude larger than any previous work.

Our research aims to answer the following questions: How much additional cost are we incurring for marginal increases in accuracy? What are the contributing factors to energy consumption in AI models? Do current acceleration techniques enhance energy efficiency? How to trade-off between energy consumption and accuracy?

To address these challenges, out key contributions are as follows:
\begin{itemize}
    \item Extensive Dataset: We provide a comprehensive dataset of inference energy consumption metrics for 1,200 ImageNet classification models, enabling better understanding and comparison of model efficiencies.
    \item Analysis of Diminishing Returns: We demonstrate the diminishing returns in accuracy improvements relative to increases in energy consumption, highlighting the need to reevaluate the pursuit of marginal accuracy gains at significantly higher energy costs.
    \item Insights into Energy Consumption Factors: We identify factors contributing to energy consumption, correct existing misunderstandings, and assess the effectiveness of improving throughput on energy consumption.
    \item Energy Efficiency Scoring System: We introduce a scoring system to rank models based on their energy efficiency, providing a standardized metric for evaluation and informed decision-making.
    \item Interactive Web Application: We develop an interactive web app that allows users to visualize and compare models based on energy efficiency and other metrics, promoting energy-conscious choices within the community.
\end{itemize}

\section{Related Work}
Several researchers have called on the AI community to raise awareness about the energy consumption of AI models and their subsequent environmental consequences. Schwartz et al.~\cite{schwartz2020green} proposed the concept of Green AI, which emphasizes computational efficiency alongside model quality, as opposed to Red AI that prioritizes higher accuracy regardless of computational cost - a norm in the field. They also advocated for reporting a model's FLOP count as a standard practice in publications. Van Wynsberghe~\cite{van2021sustainable} introduced the Sustainable AI movement to promote changes throughout the AI life cycle - including training, fine-tuning, implementation, and governance - towards ecological integrity and sustainable development.

Li et al.\cite{li2016evaluating} were among the first to investigate energy efficiency on GPUs, testing both training and inference of AlexNet, OverFeat, VGG, and GoogleNet on NVIDIA K20m and TITAN X GPUs. They identified the energy consumption of different CNN layers and analyzed the impact of hardware settings such as batch size, hyper-threading, ECC, and DVFS on energy efficiency. 

Canziani et al.\cite{canziani2016analysis} conducted a comparative analysis of over a dozen models, including variants of Inception, VGG, and ResNet, evaluating metrics such as accuracy, memory usage, inference time, and power consumption on an NVIDIA Jetson TX1. Their work aimed to guide efficient DNN design for practical applications by highlighting the trade-offs between accuracy and computational requirements. Yao et al.~\cite{yao2021evaluating} tested three CNNs - VGG16, ResNet50, and Inception-V3 - on three GPUs: NVIDIA Tesla M40, P4, and V100. They highlighted the impact of different configurations and optimizations, including quantization and the use of TensorRT and Tensor Cores, on energy consumption, providing insights for more energy-efficient deployment of CNNs in high-performance computing environments. Overall, the number of models evaluated in these works is limited, and the GPUs used are outdated by today's standards.

Henderson et al.~\cite{henderson2020towards} proposed a standardized framework for consistent reporting of energy and carbon emissions in ML research, aiming to raise awareness, enable cost-benefit analyses, and promote energy-efficient practices in model development and deployment. One of their main arguments was that a model's parameter count and FLOPs do not necessarily correlate with energy consumption. They tested over 20 models, including VGG, ResNet, MobileNet, and SqueezeNet. However, they did not specify which GPU they used. Most importantly, they ran all models with a batch size of one, which underutilizes any reasonably modern GPU and gives larger models an unfair advantage.

Desislavov et al.~\cite{desislavov2023trends} conducted one of the most extensive analyses to date, examining 94 different ImageNet classification models. They showed that efficiency gains from hardware advancements and algorithmic improvements mitigate energy growth despite increasing model complexity. However, they estimated the energy consumption of a model by dividing the model's FLOPs by the GPU's FLOPs per second and multiplying by the GPU's TDP. This is a highly idealized and optimistic assumption, which, as we show in our results section, differs significantly from real-world scenarios.

Shifting focus away from computer vision, Samsi et al.\cite{samsi2023words} benchmarked the inference energy and compute requirements of various configurations of the LLaMA model across GPU setups to highlight energy usage patterns and identify optimization opportunities for resource efficiency. Luccioni et al.\cite{luccioni2024power} benchmarked 80 models across 10 specific tasks and 8 general-purpose models, providing a systematic comparison of energy consumption and carbon emissions in AI model deployment. They emphasized the significantly higher costs of deploying general-purpose models compared to task-specific ones and urged careful consideration of these environmental impacts.

\section{Methodology \& Experimental Setup}

\paragraph{Model Selection} To perform a comprehensive analysis of energy efficiency across diverse model architectures, we included all available pretrained models from the Hugging Face PyTorch Image Models (Timm) library~\cite{rw2019timm}. Timm is a widely used repository offering the largest collection of state-of-the-art vision models, including convolutional neural networks (CNNs), vision transformers, and hybrid architectures. Leveraging this extensive collection ensures our evaluation covers a broad spectrum of model sizes, complexities, and design philosophies.

\vspace{-8pt}
\paragraph{Hardware Configuration} All experiments were conducted on two NVIDIA GPUs: the A100 PCIe 40GB~\cite{A100WhitePaper} and the H100 PCIe 80GB~\cite{H100WhitePaper}. The A100 is NVIDIA's flagship GPU from the previous "Ampere" generation, while the H100 represents the latest "Hopper" generation. Both GPUs are recognized for their state-of-the-art performance and efficiency in deep learning computations. Key hardware and software configurations are detailed in Table~\ref{tab:setup}.

\begin{table}[h]
  \centering
  \begin{tabular}{lcc}
    \toprule
     & A100 PCIe 40G  & H100 PCIe 80G\\
    \midrule
    TDP & 250W & 310W \\
    CPU & EPYC 7452 & Xeon Gold 6342\\
    RAM & 1TB & 512GB\\
    OEM & GIGABYTE & DELL?\\
    \midrule
    OS & CentOS 8.1.1911 & CentOS 8.1.1911 \\
    GPU Driver & 525.116.04 & 525.116.04 \\
    CUDA vers. & 11.8 & 11.8 \\
    PyTorch vers.  & 2.4 & 2.4 \\
    \bottomrule
  \end{tabular}
  \vspace{0.2cm}
  \caption{Key hardware and software configurations.}
  \label{tab:setup}
\end{table}

\vspace{-8pt}
\paragraph{Inference Deployment Methods} We evaluated the models using two inference methods: standard PyTorch~\cite{Ansel_PyTorch_2_Faster_2024} inference and NVIDIA's TensorRT. The standard PyTorch inference serves as a baseline without additional optimizations. TensorRT is widely adopted in industry for optimizing deep learning model deployment on NVIDIA GPUs, reflecting real-world production scenarios. By comparing these methods, we assess the impact of inference optimizations on energy consumption and performance metrics.

\vspace{-8pt}
\paragraph{Accuracy Metrics} To comprehensively assess the models' accuracy, robustness, and generalization capabilities, we utilized six validation/test datasets: the original ImageNet validation set~\cite{Imagenet-Original}, as well as 5 other well known datasets widely used in the field: ImageNet Real Labels~\cite{Imagenet-real}, ImageNet V2 Matched Frequency~\cite{Imagenet-V2}, ImageNet Sketch~\cite{Imagenet-Sketch}, ImageNet Adversarial~\cite{Imagenet-A}, and ImageNet Rendition~\cite{Imagenet-R}. These datasets evaluate not only standard classification accuracy but also the models' performance across different visual domains and distribution shifts.

\begin{figure}[h]
  \centering
  \includegraphics[width=0.4\linewidth]{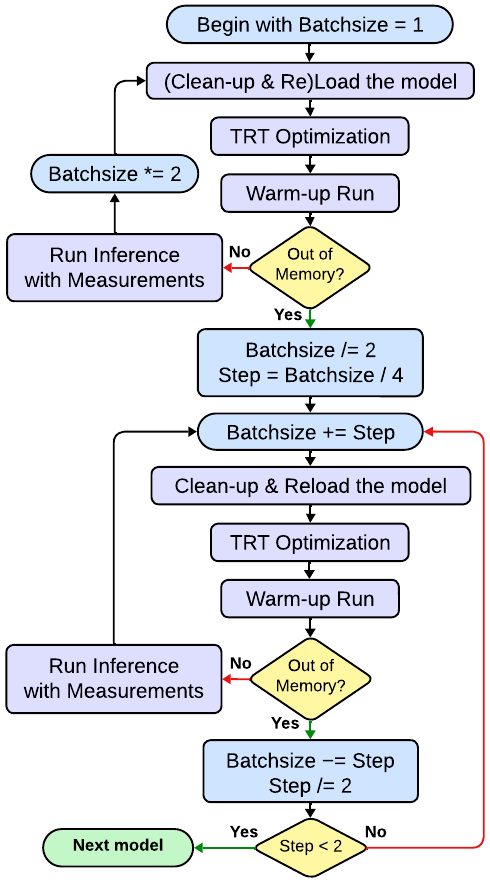}
  \vspace{-0.2cm}
   \caption{Automated model testing procedure.}
   \label{fig:exp_flowchart}
   \vspace{-0.4cm}
\end{figure}

\vspace{-8pt}
\paragraph{Measurement Procedure} We developed an automated script to test all selected models. For each model, we iteratively increased the batch size, starting from 1 and doubling until reaching the maximum size that fits into the GPU memory. Fig.\ref{fig:exp_flowchart} illustrates the procedure for each batch size. For each batch size, we performed two runs: a warm-up run that helps handle potential CUDA out-of-memory errors and warms up the GPU, followed by a measured run.

We conducted inference runs until both conditions were met: more than 13 repetitions and a runtime exceeding 10 seconds. This ensured sufficient data collection for both large and small models. Synthetic input data (random tensors) were used to eliminate I/O bottlenecks. We enforced synchronous execution to have precise control over execution timing. Timestamps were recorded after each batch to align runtime with GPU power data and to calculate throughput and latency.

\vspace{-8pt}
\paragraph{Energy Measurement} Energy consumption was measured using the GPU's onboard power sensors via nvidia-smi, following guidelines from recent work assessing its accuracy~\cite{yang2023part}. We instructed nvidia-smi to record power usage and other GPU metrics at a rate of 100 Hz, and the data were logged for analysis.

\vspace{-8pt}
\paragraph{Additional Metrics} We collected key model statistics, including the number of parameters and FLOPs, using the ptflops\cite{ptflops} and torchinfo\cite{torchinfo} libraries. Computational performance metrics such as GPU utilization, VRAM usage, and temperature were also gathered to investigate relationships between model characteristics, performance metrics, and energy consumption.

\vspace{-8pt}
\paragraph{Result Integrity} To ensure the integrity of our results, we had exclusive access to the machines during experimentation, preventing interference from other processes. All experiments were conducted on the same A100 and H100 GPU cards to maintain consistency. The servers were housed in a data center with controlled cooling, and GPU temperatures were monitored to remain within operational ranges.

\vspace{-8pt}
\paragraph{Reproducibility} The source code for our experiments is available on GitHub\footnote{\url{https://github.com/JimZeyuYang/DL-Inference-Energy-Efficiency.git}} to facilitate replication and verification of our results.

\section{Results}

In this section, we present our findings in four parts. First, we provide the overall energy consumption data of all tested models, analyzing observations and trends. Next, we investigate the factors contributing to energy consumption, aiming to correct common misunderstandings. We then explore methods to improve energy efficiency and examine the relationship between energy consumption, throughput, and the Thermal Design Power (TDP) of GPUs. Finally, we discuss the trade-off between accuracy and energy consumption and introduce our interactive web application designed to facilitate this analysis.

\begin{figure}[h!]
  \centering
  \includegraphics[width=0.6\linewidth]{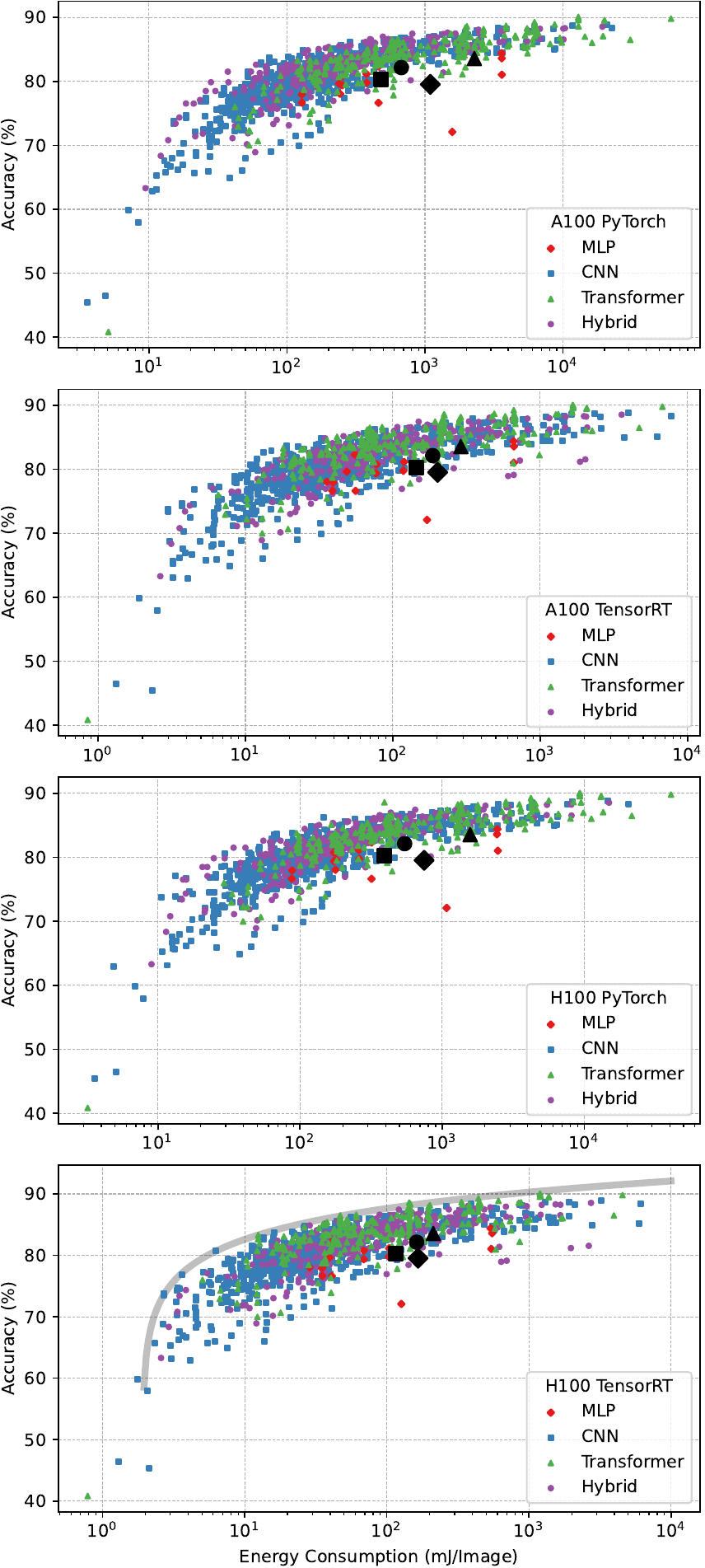}
   \caption{Energy consumption data of all tested models for the four different setups.}
   \label{fig:overall}
   \vspace{-0.8cm}
\end{figure}

\subsection{Energy Consumption Results and Trends}

\paragraph{Overall Results} Fig.\ref{fig:overall} displays the energy consumption data of all tested models under the four different inference setups. These models are categorized by their architecture into Multi-Layer Perceptrons (MLPs), Convolutional Neural Networks (CNNs), Transformers, and hybrid CNN-Transformer models. Note that the energy consumption axis is on a logarithmic scale.

A prominent observation is the steep diminishing returns in accuracy as energy consumption increases. The range of energy consumption spans approximately four orders of magnitude (a factor of 10,000). In the first decade, a tenfold increase in energy consumption leads to roughly double the accuracy, from 40\% to 80\%. The second decade increases accuracy by about 7\% (to 87\%), and a further tenfold increase raises accuracy by only 3\% (to 90\%), with negligible improvements in the last decade. This trend exhibits logarithmic growth on a logarithmic scale, indicating a nested logarithmic relationship. We fitted this growth function to the models on the efficient Pareto front of the H100 TensorRT data, shown as the grey line in Fig.\ref{fig:overall}. Extrapolating this trend suggests that a model capable of achieving 100\% accuracy would require consuming 207 MWh of electricity to classify a single image.

Comparing the different model architectures, MLP models generally have lower accuracy and higher energy consumption than the others. CNN models occupy the lower accuracy and energy consumption region, while Transformer models are situated in the high accuracy and high energy consumption region. Hybrid CNN-Transformer models are distributed between these two extremes. This relationship is more evident when comparing the centers of the clusters, represented by the large black markers in the plot.

The distribution of models remains largely consistent across the different inference setups, as shown by the correlations in Table~\ref{tab:coor_bet_setup}. Given the similarity in distributions, we will mainly use the A100 TensorRT setup as a representative for some of the subsequent analyses.

\begin{table}[h]
  \centering
  \begin{tabular}{l|cc}
    PCC / $\rho$ & A100 - PT & H100 - TRT \\
    \midrule
    A100 - TRT & 0.8553  / 0.9345 & 0.9840 / 0.9943 \\
    H100 - PT  & 0.9939  / 0.9904 & 0.8299 / 0.9201 \\
  \end{tabular}
  \vspace{0.2cm}
  \caption{Pearson Correlation Coefficient (PCC) and Spearman's Rank Correlation Coefficient ($\rho$) of the energy consumption of models across different deployment setups: comparing same GPU with different software and different GPU with same software.}
  \label{tab:coor_bet_setup}
\end{table}

\begin{figure}[h]
  \centering
  \includegraphics[width=0.6\linewidth]{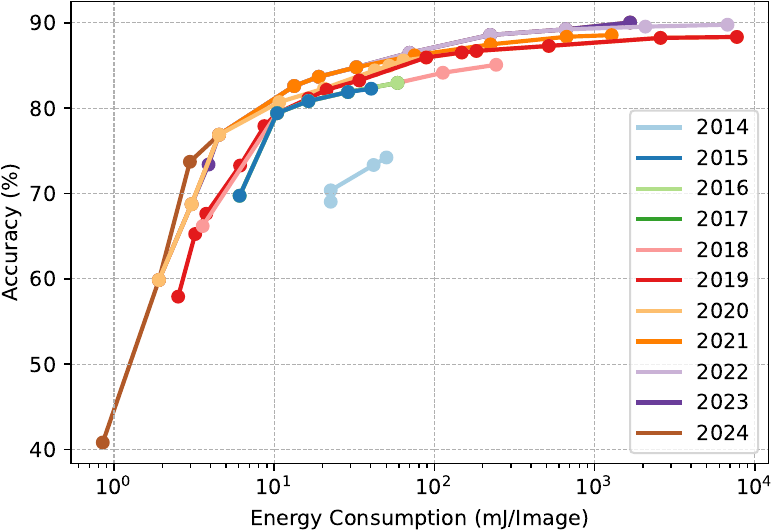}
   \caption{Yearly progress on model efficiency (A100 with TensorRT as an example for illustration).}
   \label{fig:yearly}
   \vspace{-0.4cm}
\end{figure}

\vspace{-8pt}
\paragraph{Year-on-year Improvement} We categorized the models by the year their corresponding papers were published. Fig.\ref{fig:yearly} illustrates how each year's new models push the existing convex hull of models towards greater accuracy and efficiency. As the field evolved and more models were introduced, the efficient frontier expanded in both the high accuracy, high energy consumption direction and the low accuracy, low energy consumption direction. Notably, we observe a consistent vertical increase in accuracy in the high consumption region and a somewhat inconsistent horizontal shift towards lower consumption in the lower accuracy region. However, improvements in the middle region—towards the top-left corner representing high accuracy and low consumption—appear to be more stagnant compared to others.

\subsection{Understanding of Energy consumption}
We investigate the relationship between energy consumption and various model metrics, including the number of parameters, FLOPs, activations, and input image size. Additionally, we aim to correct some misconceptions about the energy consumption of models.

\begin{figure}[h]
  \centering
  \includegraphics[width=0.6\linewidth]{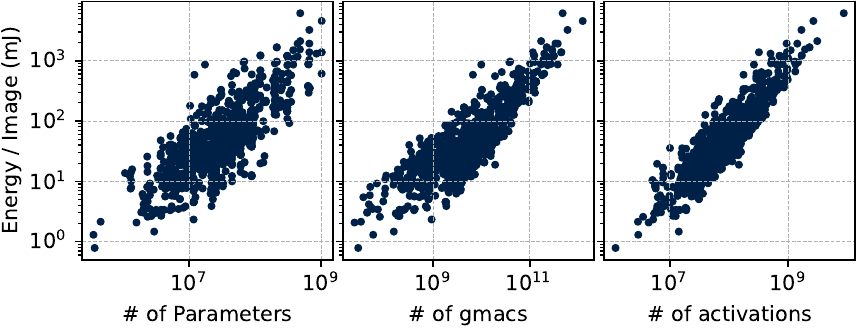}
   \caption{Relationship between energy consumption per image and number of parameters, FLOPs, and activations (A100 with TensorRT). Pearson Correlation Coefficients are 0.6572, 0.8683, and 0.8999, respectively.}
   \label{fig:PFA}
   \vspace{-0.2cm}
\end{figure}

\vspace{-8pt}
\paragraph{Parameters, FLOPs, and Activations} Intuitively, as model size and complexity increase, computational intensity grows, leading to higher energy costs. However, Henderson et al.~\cite{henderson2020towards} concluded that "FPOs and Params have no strong correlation with Energy Consumption." Contrarily, our results demonstrate that the number of parameters has a moderately strong linear correlation with energy consumption, while FLOPs and activations exhibit a very strong correlation, as shown in Fig.\ref{fig:PFA}.

\begin{figure}[h]
  \centering
  \includegraphics[width=0.6\linewidth]{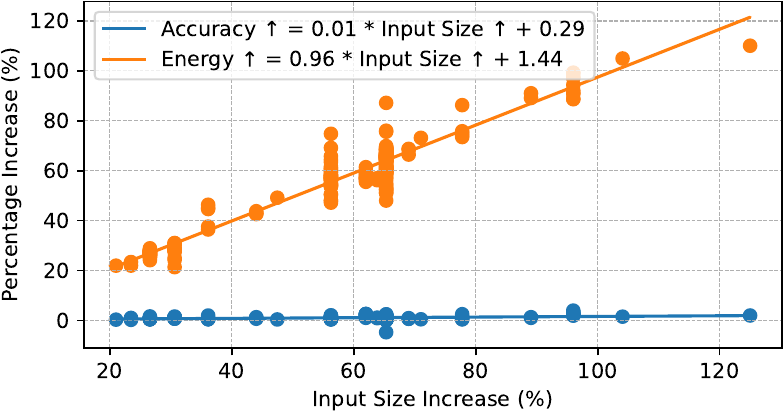}
   \caption{Increase in accuracy and energy consumption as the input image size increases (A100 with TensorRT). The increase in accuracy is minimal, whereas energy consumption is almost directly proportional with input size.}
   \label{fig:input_size}
   \vspace{-0.4cm}
\end{figure}

\vspace{-8pt}
\paragraph{Input Image Size} For models that accept variable input sizes, increasing the input size yields only marginal improvements in accuracy but results in a substantial increase in energy consumption. Fig.\ref{fig:input_size} illustrates the increase in accuracy and energy consumption as the input size increases for a subset of models supporting variable input sizes.

\begin{figure}[h]
  \centering
  \includegraphics[width=0.6\linewidth]{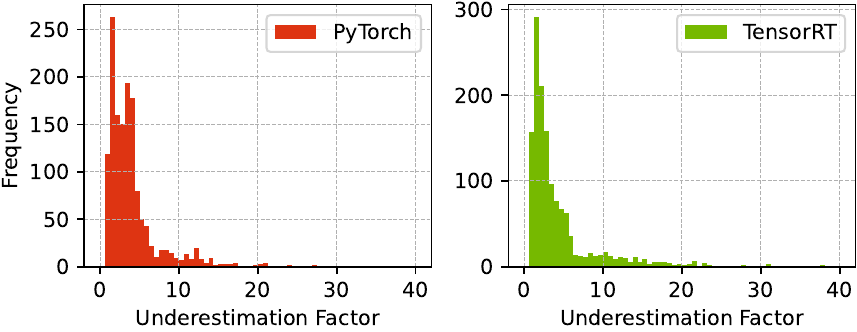}
   \caption{Error in naive estimation of energy consumption based on FLOPs compared with actual measurements on the A100 GPU. The geometric mean of the underestimation factor is 3.13 and 3.16, respectively.}
   \label{fig:estimate}
   \vspace{-0.4cm}
\end{figure}

\vspace{-8pt}
\paragraph{Error of Naive Energy Estimation} As mentioned in the related work section, Desislavov et al.~\cite{desislavov2023trends} estimated the energy consumption of a model by dividing the model's FLOPs by the GPU's FLOPs per second and multiplying by the GPU's TDP. We replicated this calculation and compared the estimated energy consumption with our actual measurements. Fig.\ref{fig:estimate} shows the distribution of underestimation. On average, this method underestimates energy consumption by approximately three times and can underestimate by nearly 40 times in some cases.

\subsection{Improving energy efficiency}
\label{sec:TDP}

\begin{figure}[h]
  \centering
  \includegraphics[width=0.6\linewidth]{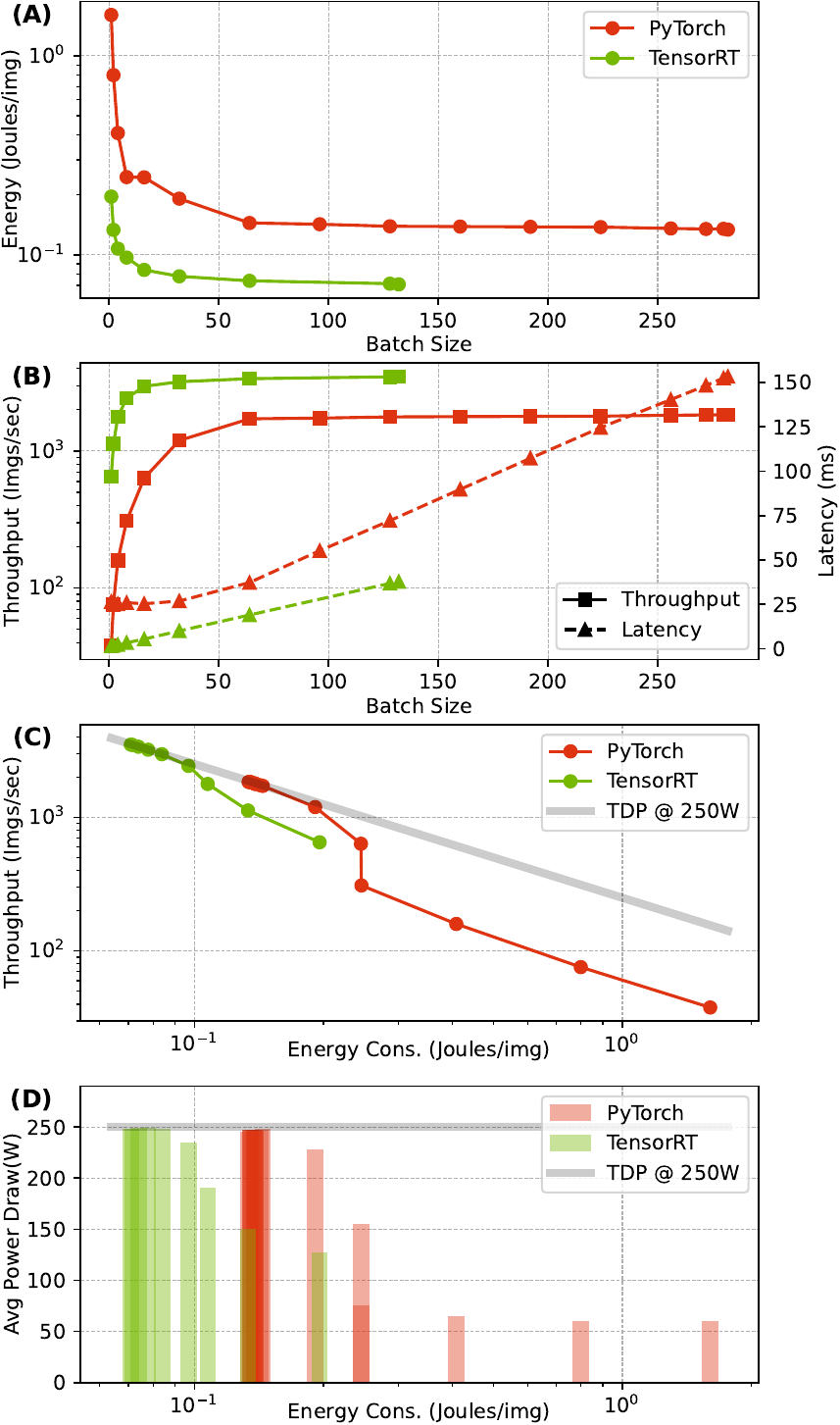}
   \caption{Analysis of batch size effects on EfficientViT on H100: (A) Energy consumption at different batch sizes, (B) Throughput and latency, (C) Throughput vs energy consumption with the TDP limit indicated, and (D) Average power draw relative to batch size.}
   \label{fig:BS_analysis}
\end{figure}

\paragraph{Batch size \& GPU Utilization} We examine the effect of batch size on inference energy consumption. Larger batch sizes enable parallel processing of multiple inputs, improving hardware utilization and reducing per-sample overhead, thereby increasing efficiency. Fig.\ref{fig:BS_analysis}A shows the energy consumption of EfficientViT on the A100 GPU with increasing batch sizes up to the GPU memory limit. As batch size increases, energy consumption decreases significantly until the memory bandwidth and processing units become fully saturated. Beyond this point, the decrease in energy consumption becomes marginal. In extreme cases, to accommodate a large batch size, CUDA libraries may select less efficient strategies that save memory footprint but increase energy consumption.

\vspace{-8pt}
\paragraph{Energy consumption, Throughput, and TDP} Fig.\ref{fig:BS_analysis}B shows the throughput and latency achieved at varying batch sizes. Latency increases linearly with batch size, while throughput increases initially and then plateaus. We observed an inverse relationship between throughput and energy consumption. Throughput is measured in images per second, and energy consumption is in joules per image. The product of these two gives the average power draw of the GPU during model execution:
\[ \frac{Imgs}{s} \times \frac{Joules}{Img} = \frac{Joules}{s} = Watt = Avg Pwr Draw\]

The maximum power a GPU can draw is its TDP, and the GPU maintains power draw not to exceed the TDP under heavy load.

To investigate this inverse relationship, we plotted throughput against energy consumption in Fig.\ref{fig:BS_analysis}C, along with the maximum possible combination of the two (the TDP line). As batch size increases, energy consumption decreases while throughput increases. Although the relationship is not strictly linear, when batch size is small, the GPU is underutilized, resulting in an average power draw below the TDP. As batch size increases, GPU utilization improves, and the throughput-energy consumption combination approaches the TDP limit. The conclusion, although counterintuitive, is that higher average power draw leads to greater energy efficiency due to better GPU utilization.

This observation underscores the importance of iterating through various batch sizes to find the most efficient configuration for a particular model on a specific GPU, ensuring a fair comparison between different models. Simply performing inference with a batch size of one would unfairly favor larger and more complex models, as they naturally consume more GPU resources.

\begin{figure}[h]
  \centering
  \includegraphics[width=0.6\linewidth]{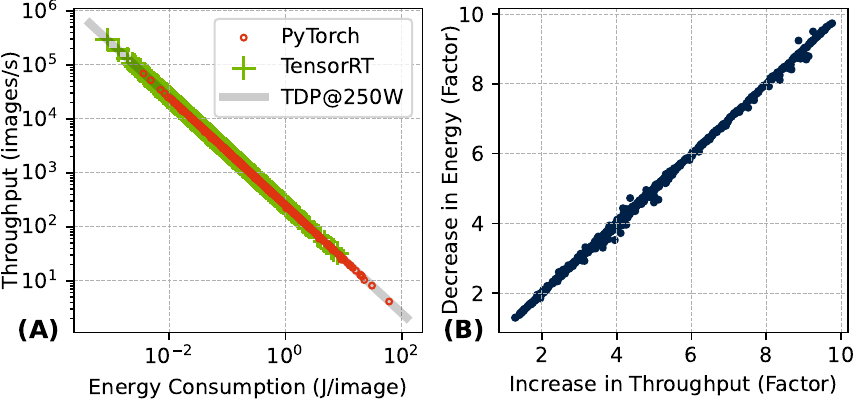}
   \caption{Relationship between energy consumption, throughput, and TDP: (A) Energy consumption versus throughput for all models, showing the TDP limit, and (B) Improvement in energy consumption versus throughput when using TensorRT instead of PyTorch on the A100 GPU.}
   \label{fig:TDP_Throughput}
\end{figure}

Fig.\ref{fig:TDP_Throughput}A shows the energy consumption and throughput of all models at their most efficient batch size on the A100 GPU using PyTorch and TensorRT. All models lie on the TDP line. Fig.\ref{fig:TDP_Throughput}B illustrates the improvement in energy consumption versus throughput when switching from PyTorch to TensorRT. There is a near-perfect directly proportional relationship between the increase in throughput and the decrease in energy consumption. This is encouraging, as many existing inference acceleration methods target throughput, and the TDP ceiling ensures that any gain in throughput must come from a decrease in energy consumption.

\begin{figure}[t]
  \centering
  \includegraphics[width=0.6\linewidth]{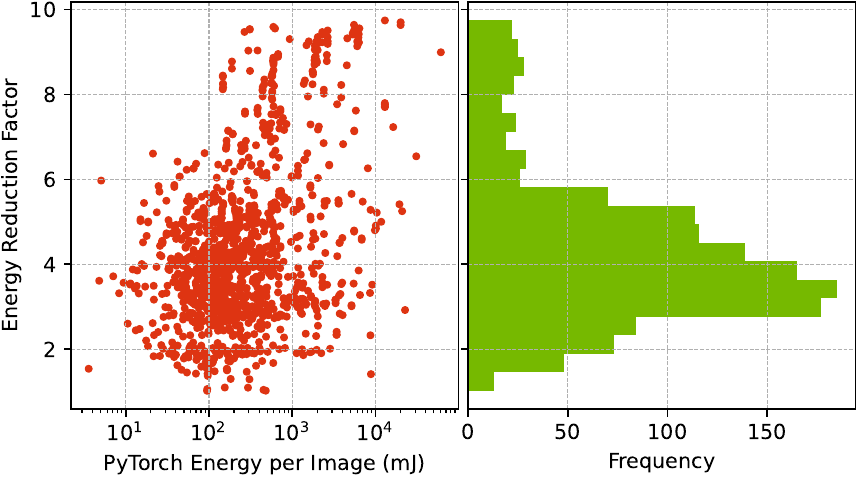}
   \caption{Reduction in energy consumption when using TensorRT instead of PyTorch on the A100. The geometric mean reduction factor is 3.89 with a geometric standard deviation of 1.53. On the H100, the geometric mean reduction factor is 4.02 with a geometric standard deviation of 1.55.}
   \label{fig:TRT_reduction}
\end{figure}

\vspace{-8pt}
\paragraph{Energy Savings from TensorRT} Fig.\ref{fig:TRT_reduction} shows the reduction in energy consumption for each model when switching from PyTorch to TensorRT. On average, energy consumption is reduced by approximately four times. Notably, models that were highly energy-consuming in PyTorch achieve more significant reductions (up to 10$\times$) when using TensorRT, presumably because the PyTorch implementation is less efficient. This also explains why the energy consumption correlation between PyTorch and TensorRT on the same GPU, as shown in the previous section, was relatively low compared using the same software on different GPUs.

\begin{figure}[ht]
    \centering
    \includegraphics[width=\textwidth]{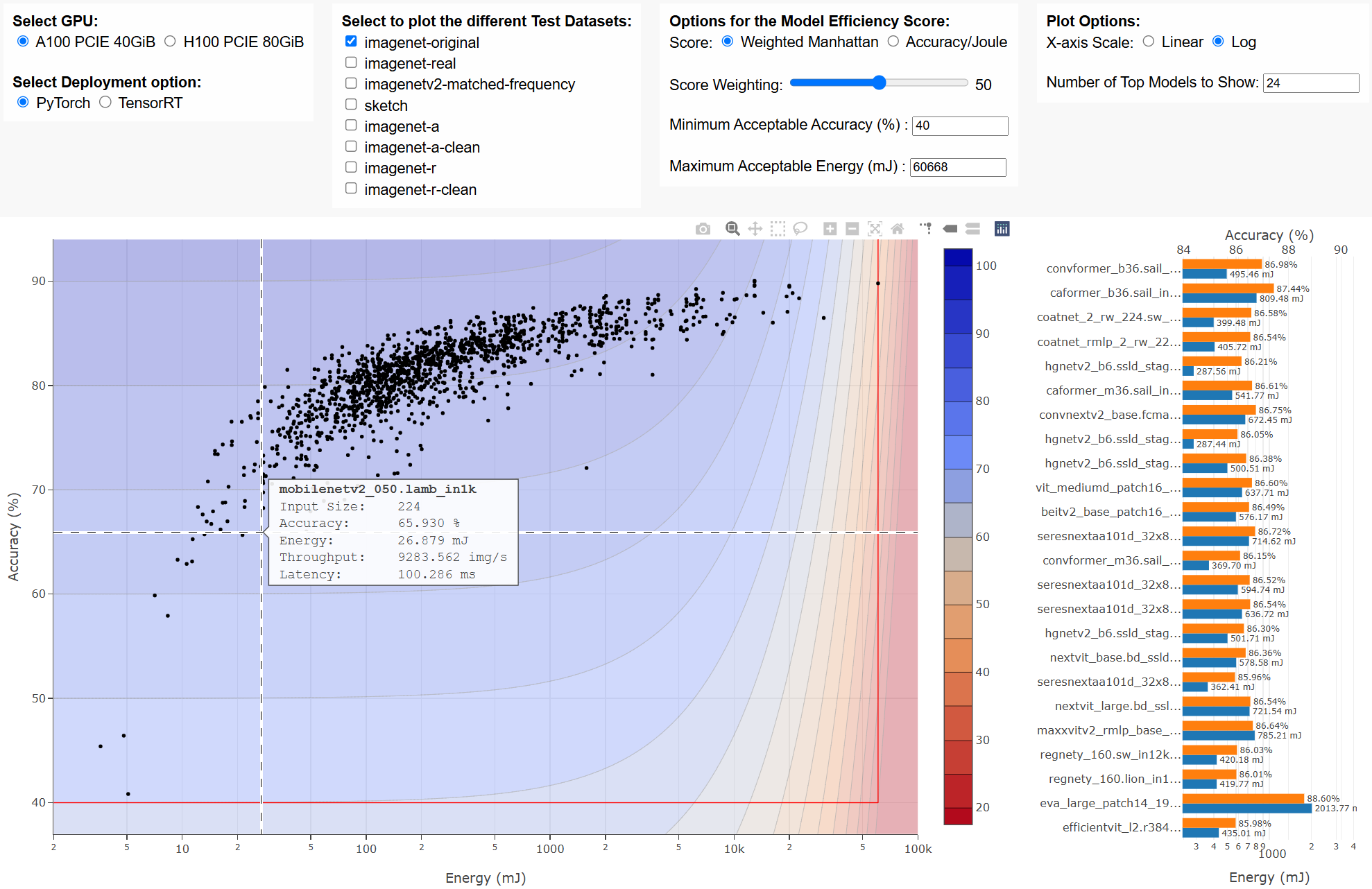}
    \caption{Screenshot of the interactive web application. The top menu allows users to select inference setups, test datasets, scoring metrics, and plotting options. The scatter plot displays energy consumption versus accuracy for all models based on the selected parameters. Hovering over a point reveals the model's details, and clicking it opens the corresponding Hugging Face webpage. The red lines indicate the accuracy and energy thresholds set by the user, and the background shows the contour map of the selected efficiency score. The bar plot on the right highlights the top-performing models under the chosen scoring metric. The web application is available at: \url{https://jimzeyuyang.github.io/DL-Inference-Energy-Efficiency/}.}
    \label{fig:webpage}
    \vspace{-0.6cm}
\end{figure}

\subsection{Trade off between Energy Consumption and Model Accuracy}
\label{sec:result_tradeoff}
We propose two methods to evaluate the trade-off between energy consumption and achieved accuracy. The standard way of calculating efficiency is the ratio of the desired output to the resources consumed, often referred to as "bang for the buck." In this context, the efficiency measure would be the model accuracy divided by the energy consumption, with units of percentage per joule.

However, this metric may unfairly favor less accurate models. For example, a trivial model that always outputs ``goldfish" might achieve 0.1\% accuracy on ImageNet-1K but consume negligible energy, resulting in a deceptively high efficiency score despite being practically useless.

To mitigate this issue, we suggest using this measure in conjunction with a minimum accuracy threshold. This approach allows us to rank models based on energy efficiency while ensuring they meet a baseline level of accuracy.

Additionally, similar to prioritizing accuracy, we may not always prefer maximum efficiency at all costs. In critical applications like autonomous driving or medical diagnostics, accuracy may be prioritized over energy consumption. Therefore, we propose a metric that uses a weighted Manhattan distance between the model's performance and the ideal point (100\% accuracy and 0 energy consumption):
\[ score = 100 - \left( W \left(\frac{E}{N}\right) + \left(1-W\right) \left(100-A\right) \right) \]

where $E$ is the energy consumption, $A$ is the accuracy, $N$ is a normalization term equal to the maximum energy consumption among the models (to normalize the score within the same range), and $W$ is a weight between 0 and 1. A weight of 1 indicates complete emphasis on energy consumption, disregarding accuracy, and vice versa for a weight of 0. Users can adjust this weight to balance their specific needs between accuracy and energy consumption.

To facilitate the exploration of the trade-off between energy consumption and model accuracy, we have developed an interactive web application. This tool allows users to visualize and compare the energy efficiency and accuracy of all the models in our dataset under various configurations. Users can adjust parameters such as the inference setup (GPU and software library), select specific or mean of multiple test datasets, set minimum acceptable thresholds for accuracy and energy consumption, and apply different scoring metrics. The application provides real-time updates to the plots based on user selections, aiding in informed decision-making for model selection. A screenshot of the webpage is shown in Fig.\ref{fig:webpage}.

\section{Discussion}

Our extensive benchmarking shows that while state-of-the-art models achieve higher accuracy, they incur significant energy costs with diminishing returns. The logarithmic increase in energy consumption for minimal accuracy gains raises concerns about the sustainability of this trend, given environmental and economic impacts.

Practitioners face trade-offs among accuracy, throughput, and energy consumption—the "iron triangle." Our findings in Section~\ref{sec:TDP} illustrate that throughput and energy consumption are constrained by the GPU's TDP, highlighting the importance of optimizing models for efficiency. The interactive tool introduced in Section~\ref{sec:result_tradeoff} enables users to visualize and balance these trade-offs, facilitating informed decision-making.

Our study significantly extends previous research by measuring the inference energy consumption of 1,200 models on cutting-edge GPUs, and unlike works that relied on estimations or outdated hardware, we ensured a fair comparison between the models, providing a comprehensive and accurate understanding. While our experiments focus on A100 and H100 GPUs, we believe the observed trends are applicable to other GPUs, as we showed that power consumption has a high correlation across different GPUs.

The significant energy costs associated with marginal accuracy gains underscore the need for a shift toward more sustainable AI practices. By emphasizing efficiency and providing tools to evaluate energy consumption, our work encourages the development and adoption of models that balance performance with environmental impact, aligning with the growing movement toward Sustainable AI~\cite{van2021sustainable}.

We plan to continuously update our dataset and interactive web application as new models and GPUs become available. We encourage the community to use our open-source code to measure the energy consumption of additional models and submit their results. By collaborating, we can build a more comprehensive resource that tracks energy efficiency trends over time, fosters transparency, and accelerates the development of sustainable AI technologies.

\section{Conclusion}
In summary, our work provides a comprehensive analysis of the energy consumption of 1200 vision models, illuminating the significant trade-offs between model accuracy and energy efficiency. By highlighting the diminishing returns in accuracy gains and introducing practical tools and metrics, we hope to shift the focus towards more sustainable AI practices. Our findings lay the groundwork for further exploration into optimizing deep learning models for energy efficiency, encouraging a paradigm shift in how we evaluate and prioritize model performance.

\section*{Acknowledgments}
The authors would like to acknowledge the use of the University of Oxford Advanced Research Computing (ARC) facility in carrying out this work. http://dx.doi.org/10.5281/zenodo.22558

\bibliographystyle{unsrt}  
\bibliography{refs}

\end{document}